\documentclass[runningheads]{llncs}
\usepackage{graphicx}
\usepackage{amsmath,amssymb} 
\usepackage{color}
\usepackage[width=122mm,left=12mm,paperwidth=146mm,height=193mm,top=12mm,paperheight=217mm]{geometry}
\usepackage{xspace}
\usepackage{cite}
\usepackage{adjustbox}

\makeatletter
\DeclareRobustCommand\onedot{\futurelet\@let@token\@onedot}
\def\@onedot{\ifx\@let@token.\else.\null\fi\xspace}

\def\eg{\emph{e.g}\onedot} 
\def\ie{\emph{i.e}\onedot} 
 
 \def\vs{\emph{vs}\onedot}
 
\def\etal{\emph{et al}\onedot}
\makeatother

\newcommand{\norm}[1]{\left\lVert#1\right\rVert}

\newcommand{\kevin}[1]{\ifdefined\DRAFT \textcolor{red}{(kevin: {#1})} \else \fi}

\newcommand{\eat}[1]{} 

\begin{document}
\pagestyle{headings}

\mainmatter

\def\ECCV18SubNumber{1644}  

\title{PersonLab: Person Pose Estimation and Instance Segmentation with a Bottom-Up, Part-Based, Geometric Embedding Model}

\titlerunning{PersonLab: Person Pose Estimation and Instance Segmentation}

\authorrunning{Papandreou, Zhu, Chen, Gidaris, Tompson, Murphy}

\author{George Papandreou, Tyler Zhu, Liang-Chieh Chen, Spyros Gidaris,\\Jonathan Tompson, Kevin Murphy}
\institute{Google, Inc.}

\maketitle

\begin{abstract}
We present a box-free bottom-up approach for the tasks of pose estimation and instance segmentation of people in multi-person images using an efficient single-shot model. The proposed PersonLab model tackles both semantic-level reasoning and object-part associations using part-based modeling. Our model employs a convolutional network which learns to detect individual keypoints and predict their relative displacements, allowing us to group keypoints into person pose instances. Further, we propose a part-induced geometric embedding descriptor which allows us to associate semantic person pixels with their corresponding person instance, delivering instance-level person segmentations. Our system is based on a fully-convolutional architecture and allows for efficient inference, with runtime essentially independent of the number of people present in the scene. Trained on COCO data alone, our system achieves COCO test-dev keypoint average precision of 0.665 using single-scale inference and 0.687 using multi-scale inference, significantly outperforming all previous bottom-up pose estimation systems. We are also the first bottom-up method to report competitive results for the person class in the COCO instance segmentation task, achieving a person category average precision of 0.417.
\end{abstract}

\section{Introduction}
\label{sec:intro}

The rapid recent progress in computer vision has allowed the community to move beyond classic tasks such as bounding box-level face and body detection towards more detailed visual understanding of people in unconstrained environments. In this work we tackle in a unified manner the tasks of multi-person detection, 2-D pose estimation, and instance segmentation. Given a potentially cluttered and crowded `in-the-wild' image, our goal is to identify every person instance, localize its facial and body keypoints, and estimate its instance segmentation mask. A host of computer vision applications such as smart photo editing, person and activity recognition, virtual or augmented reality, and robotics can benefit from progress in these challenging tasks.

There are  two main  approaches for tackling  multi-person detection, pose estimation and segmentation.
The \emph{top-down} approach starts by identifying and roughly localizing individual person instances by means of a bounding box object detector, followed by single-person pose estimation or binary foreground/ background segmentation in the region inside the bounding box.
By contrast, the
\emph{bottom-up} approach starts by localizing identity-free semantic entities (individual keypoint proposals or semantic person segmentation labels, respectively), followed by grouping them into person instances.
In this paper, we adopt the latter approach. We develop a box-free fully convolutional system whose computational cost is essentially independent of the number of people present in the scene and only depends on the cost of the CNN feature extraction backbone.

In particular, our approach first predicts all keypoints for every person in the image in a fully convolutional way.
We also learn to predict the relative displacement between each pair of keypoints, also proposing a novel recurrent scheme which greatly improves the accuracy of long-range predictions.
Once we have localized the keypoints, we use a greedy decoding process to group them into instances. Our approach starts from the most confident detection, as opposed to always starting from a distinguished landmark such as the nose, so it works well even in clutter.

In addition to predicting the sparse keypoints, our system also predicts dense instance segmentation masks for each person. For this purpose, we train our network to predict instance-agnostic semantic person segmentation maps. For every person pixel we also predict offset vectors to each of the $K$ keypoints of the corresponding person instance. The corresponding vector fields can be thought as a geometric embedding representation and induce basins of attraction around each person instance, leading to an efficient association algorithm: For each pixel $x_i$, we predict the locations of all $K$ keypoints for the corresponding person that $x_i$ belongs to; we then compare this to all candidate detected people $j$ (in terms of average keypoint distance), weighted by the keypoint detection probability;
if this distance is low enough, we assign pixel $i$ to person $j$.

We train our model on the standard COCO keypoint dataset \cite{keypointchallenge}, which annotates multiple people with 12 body and 5 facial keypoints. We significantly outperform the best previous bottom-up approach to keypoint localization~\cite{newell2017associative}, improving the keypoint AP from 0.655 to 0.687.
In addition, we are the first bottom-up method to report competitive results on the person class for the COCO instance segmentation task. We get a mask AP of 0.417, which outperforms the strong top-down FCIS method of \cite{dai2017fully},
which gets 0.386. Furthermore our method is very simple and hence fast, since it does not require any second stage box-based refinement, or clustering algorithm. We believe it will therefore be quite useful for a variety of applications, especially since it lends itself to deployment in mobile phones.

\eat{
since it is simpler and often  faster.
It is simpler because we can detect all the parts in a fully convolutional way, without needing operations such as crop-and-resize or ROI-pooling.
It is faster since the cost of grouping the keypoints into instances is very low, as we will show.
By contrast, in the top-down approach, we have to run a separate pose prediction network on every detected bounding box; even if we reuse the 
}

\section{Related work}
\label{sec:related}

\subsection{Pose estimation}
\label{sec:related:pose}

Proir to the recent trend towards deep convolutional networks~\cite{LeCun1998, krizhevsky2012imagenet}, early successful models for human pose estimation centered around inference mechanisms on part-based graphical models~\cite{Fischler73, FelzenszwalbDPM}, representing a person by a collection of configurable parts. Following this work, many methods have been proposed to develop tractable inference algorithms for solving the energy minimization that captures rich dependencies among body parts \cite{andriluka2009pictorial, BetterAppearancePic, Sapp2010, yang11cvpr, dantone13cvpr, johnson11cvpr, pishchulin13cvpr, modec2013, armlets2013}. While the forward inference mechanism of this work differs to these early DPM-based models, we similarly propose a bottom-up approach for grouping part detections to person instances.

Recently, models based on modern large scale convolutional networks have achieved state-of-art performance on both single-person pose estimation \cite{deeppose, jainiclr2014, tompsonnips2014, Chen_NIPS14, tompson2015efficient, stackedhourglass, andriluka14cvpr, bulat2016, zisserman2016, chain16} and multi-person pose estimation \cite{deepcut, deeper_cut, insafutdinov2016articulated, iqbal2016multi, wei2016convolutional, cmu_mscoco, papandreou2017towards, he2017mask}.
Broadly speaking, there are two main approaches to pose-estimation in the literature: top-down (person first) and bottom-up (parts first).
Examples of the former include  G-RMI \cite{papandreou2017towards}, CFN \cite{huang2017coarse}, RMPE \cite{fang2017rmpe}, Mask R-CNN \cite{he2017mask}, and CPN \cite{chen2017cascaded}.
These methods all predict key point locations within person bounding boxes obtained by a person detector
(\eg, Fast-RCNN \cite{girshick2015fast}, Faster-RCNN \cite{ren2015faster} or R-FCN \cite{dai2016rfcn}).
 %

In the bottom-up approach, we first detect body parts and then group these parts to human instances. Pishchulin \etal~\cite{deepcut}, Insafutdinov \etal~\cite{deeper_cut, insafutdinov2016articulated}, and Iqbal \etal~\cite{iqbal2016multi} formulate the problem of multi-person pose estimation as part grouping and labeling via a Linear Program. Cao \etal \cite{cmu_mscoco} incorporate the unary joint detector modified from \cite{wei2016convolutional} with a part affinity field and greedily generate person instance proposals. Newell \etal~\cite{newell2017associative} propose associative embedding to identify key point detections from the same person.

\subsection{Instance segmentation}
\label{sec:related:segmentation}

The approaches for instance segmentation can also be categorized into the two top-down and bottom-up paradigms.

Top-down methods exploit state-of-art detection models to either classify mask proposals \cite{carreira2012cpmc, arbelaez2014multiscale, hariharan2014simultaneous, pinheiro2015learning, dai2015convolutional, pinheiro2016learning, dai2016instancesensitive} or to obtain mask segmentation results by refining the bounding box proposals \cite{dai2016instance, dai2017fully, he2017mask, peng2018megdet, masklab2018, liu2018path}.

Ours is a bottom-up approach, in which we associate pixel-level predictions to each object instance. Many recent models propose similar forms of instance-level bottom-up clustering. For instance, Liang \etal use a proposal-free network \cite{liang2015proposal} to cluster semantic segmentation results to obtain instance segmentation. Uhrig \etal \cite{uhrig2016pixel} first predict each pixel's direction towards its instance center and then employ template matching to decode and cluster the instance segmentation result. Zhang \etal \cite{zhang2015monocular,zhang2016instance} predict instance ID by encoding the object depth ordering within a patch and use this depth ordering to cluster instances. Wu \etal~\cite{wu2016bridging} use a prediction network followed by a Hough transform-like approach to perform prediction instance clustering. In this work, we similarly perform a Hough voting of multiple predictions. In a slightly different formulation, Liu \etal \cite{liu2016multi} segment and aggregate segmentation results from dense multi-scale patches, and aggregate localized patches into complete object instances. Levinkov \etal \cite{levinkov2017joint} formulate the instance segmentation problem as a combinatorial optimization problem that consists of graph decomposition and node labeling and propose efficient local search algorithms to iteratively refine an initial solution. InstanceCut \cite{kirillov2017instancecut} and the work of \cite{jin2016object} propose to predict object boundaries to separate instances. \cite{newell2017associative, fathi2017semantic, de2017semantic} group pixel predictions that have similar values in the learned embedding space to obtain instance segmentation results. Bai and Urtasun \cite{bai2017deep} propose a Watershed Transform Network which produces an energy map where object instances are represented as basin. Liu \etal \cite{liu2017sgn} propose the Sequential Grouping Network which decomposes the instance segmentation problem into several sub-grouping problems.

 
\section{Methods}
\label{sec:methods}

\begin{figure}[t]
    \centering
    \includegraphics[width=0.7\linewidth]{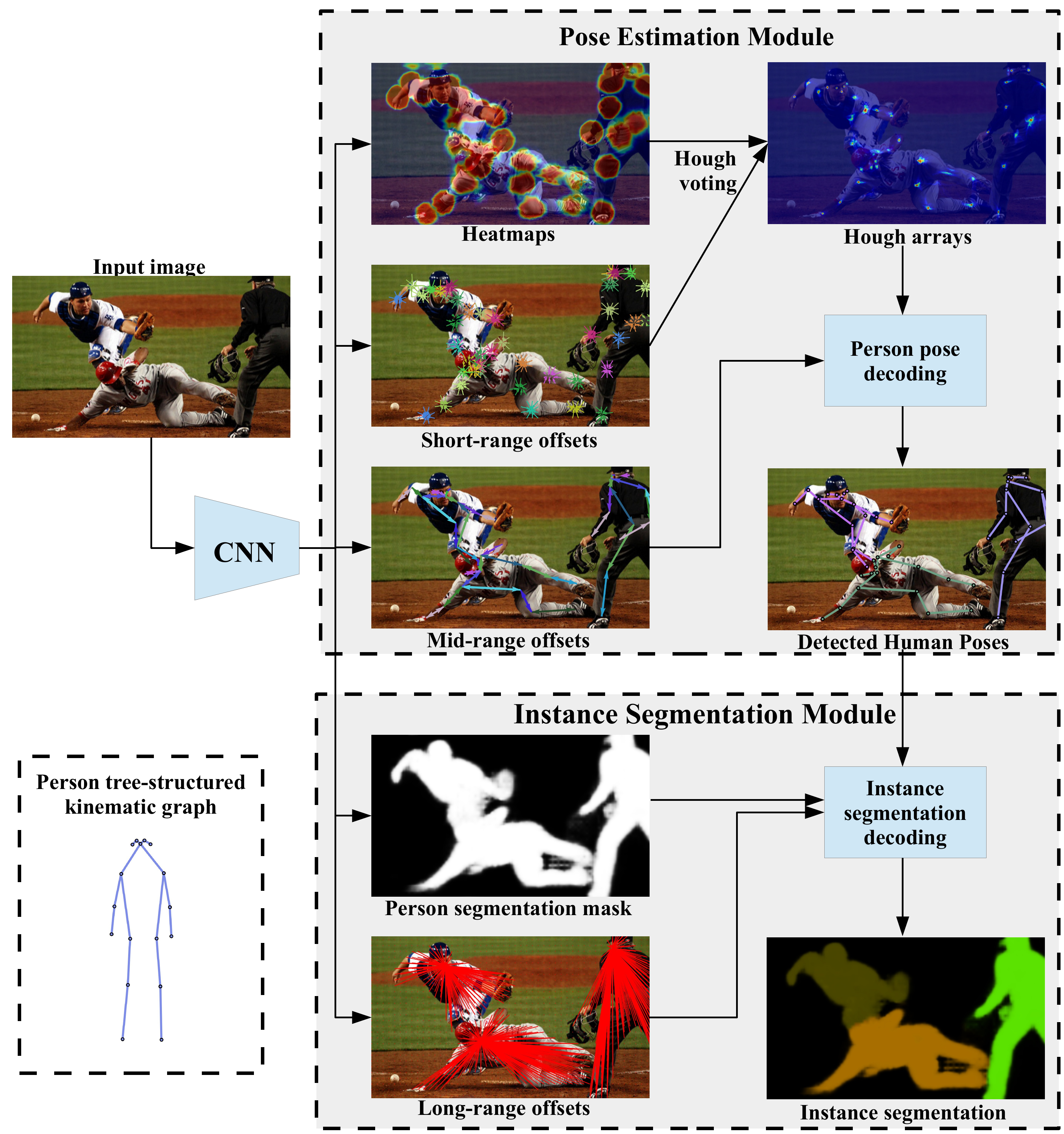}
    \vspace{-10pt}
    \caption{Our PersonLab system consists of a CNN model that predicts:
    (1) keypoint heatmaps,
    (2) short-range offsets,
    (3) mid-range pairwise offsets,
    (4) person segmentation maps, and
    (5) long-range offsets.
    The first three predictions are used by the \textit{Pose Estimation Module} in order to detect human poses while the latter two, along with the human pose detections, are used by the \textit{Instance Segmentation Module} in order to predict person instance segmentation masks.
    }
    \label{fig:pipeline}
    \vspace{-13pt}
\end{figure}

Figure~\ref{fig:pipeline} gives an overview of our system, which we describe in detail next.

\subsection{Person detection and pose estimation}

We develop a box-free bottom-up approach for person detection and pose estimation. It consists of two sequential steps, detection of $K$ keypoints, followed by grouping them into person instances. We train our network in a supervised fashion, using the ground truth annotations of the $K = 17$ face and body parts in the COCO dataset.

\subsubsection*{Keypoint detection}

The goal of this stage is to detect, in an instance-agnostic fashion, all visible keypoints belonging to any person in the image.

For this purpose, we follow the hybrid classification and regression approach of \cite{papandreou2017towards}, adapting it to our multi-person setting. We produce heatmaps (one channel per keypoint) and offsets (two channels per keypoint for displacements in the horizontal and vertical directions). Let $x_i$ be the 2-D position in the image, where $i = 1, \dots N$ is indexing the position in the image and $N$ is the number of pixels. Let $\mathcal{D}_R(y) = \{x: \norm{x - y} \le R\}$ be a disk of radius $R$ centered around $y$. Also let $y_{j,k}$ be the 2-D position of the $k$-th keypoint of the $j$-th person instance, with $j = 1, \dots, M$, where $M$ is the number of person instances in the image.

For every keypoint type $k = 1, \dots, K$, we set up a binary classification task as follows. We predict a heatmap $p_k(x)$ such that $p_k(x) = 1$ if $x \in \mathcal{D}_R(y_{j,k})$ for any person instance $j$, otherwise $p_k(x) = 0$. We thus have $K$ independent dense binary classification tasks, one for each keypoint type. Each amounts to predicting a disk of radius $R$ around a specific keypoint type of any person in the image. The disk radius value is set to $R=32$ pixels for all experiments reported in this paper and is independent of the person instance scale. We have deliberately opted for a disk radius which does not scale with the instance size in order to equally weigh all person instances in the classification loss. During training, we compute the heatmap loss as the average logistic loss along image positions and we back-propagate across the full image, only excluding areas that contain people that have not been fully annotated with keypoints (person crowd areas and small scale person segments in the COCO dataset).

In addition to the heatmaps, we also predict \emph{short-range} offset vectors $S_k(x)$ whose purpose is to improve the keypoint localization accuracy. At each position $x$ within the keypoint disks and for each keypoint type $k$, the short-range 2-D offset vector $S_k(x) = y_{j,k} - x$  points from the image position $x$ to the $k$-th keypoint of the closest person instance $j$, as illustrated in Fig.~\ref{fig:pipeline}. We generate $K$ such vector fields, solving a 2-D regression problem at each image position and keypoint independently. During training, we penalize the short-range offset prediction errors with the $L_1$ loss, averaging and back-propagating the errors only at the positions $x \in \mathcal{D}_R(y_{j,k})$ in the keypoint disks. We divide the  errors in the short-range offsets (and all other regression tasks described in the paper) by the radius $R=32$ pixels in order to normalize them and make their dynamic range commensurate with the heatmap classification loss.

We aggregate the heatmap and short-range offsets via Hough voting into 2-D Hough score maps $h_k(x), k=1, \dots, K$, using independent Hough accumulators for each keypoint type. Each image position casts a vote to each keypoint channel $k$ with weight equal to its activation probability,
\begin{equation}
h_k(x) = \frac{1}{\pi R^2}\sum_{i=1:N} p_k(x_i) B(x_i + S_k(x_i) - x) \,,
\label{eq:hough_voting}
\end{equation}
where $B(\cdot)$ denotes the bilinear interpolation kernel. The resulting highly localized Hough score maps $h_k(x)$ are illustrated in Fig.~\ref{fig:pipeline}.

\subsubsection*{Grouping keypoints into person detection instances}
\label{sec:keypoint_grouping}

\begin{figure}[t]
    \centering
    \begin{tabular}{ccc}
    \includegraphics[width=0.33\linewidth]{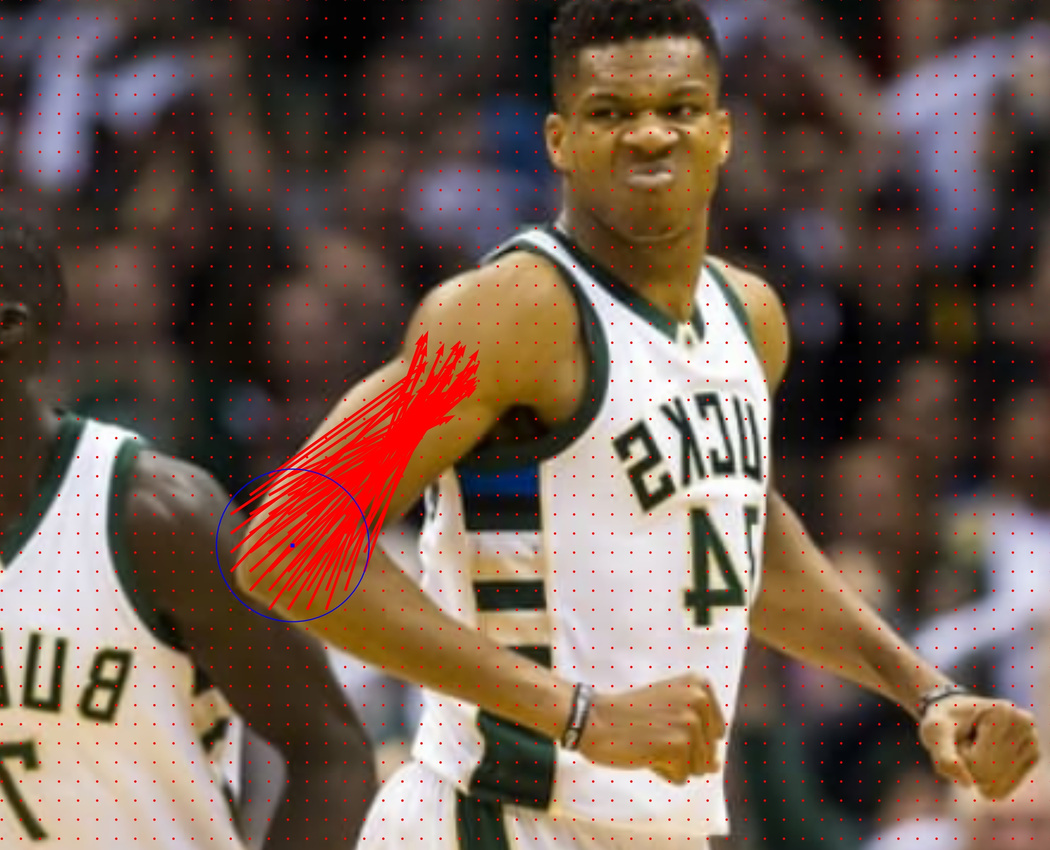} &
    \includegraphics[width=0.27\linewidth]{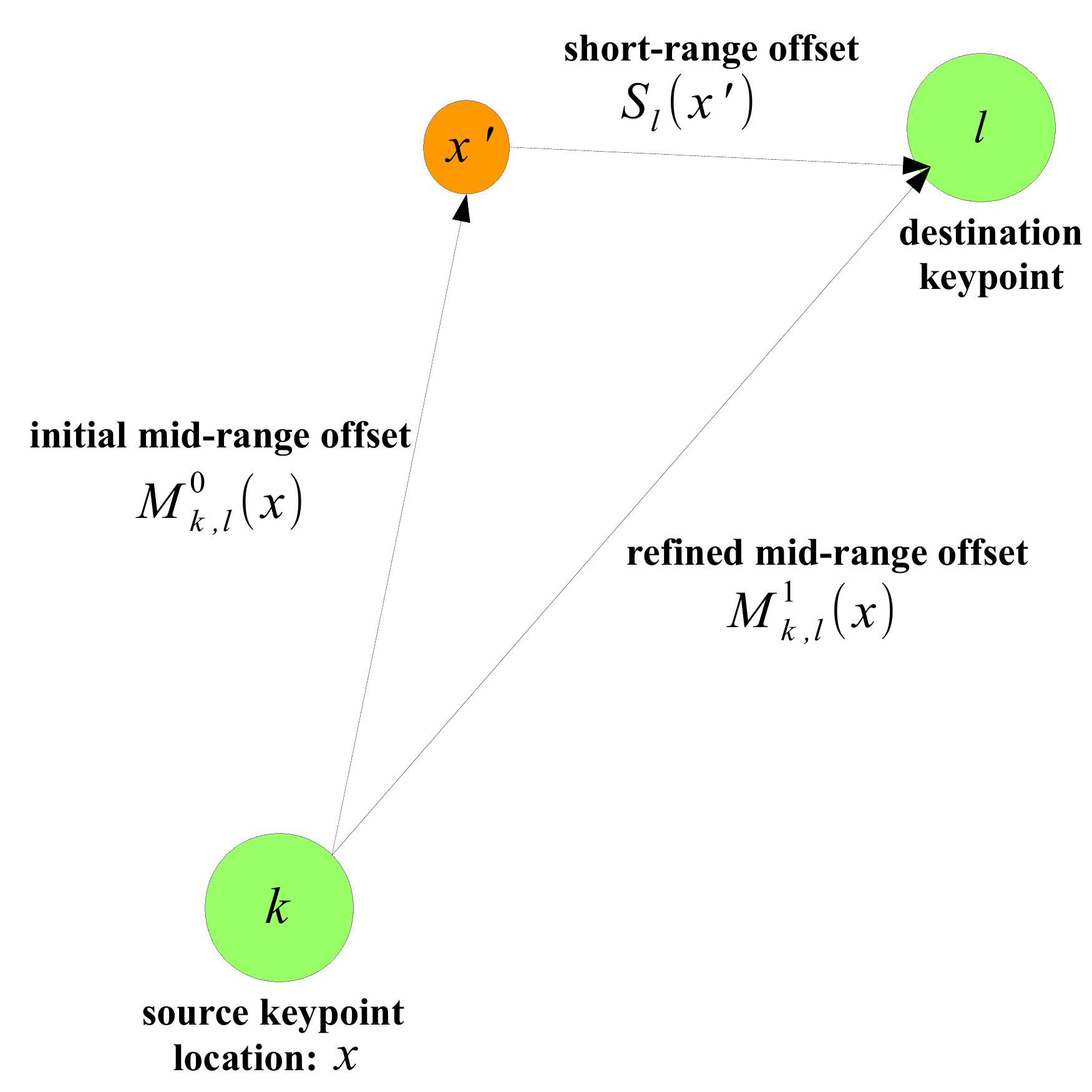} &
    \includegraphics[width=0.33\linewidth]{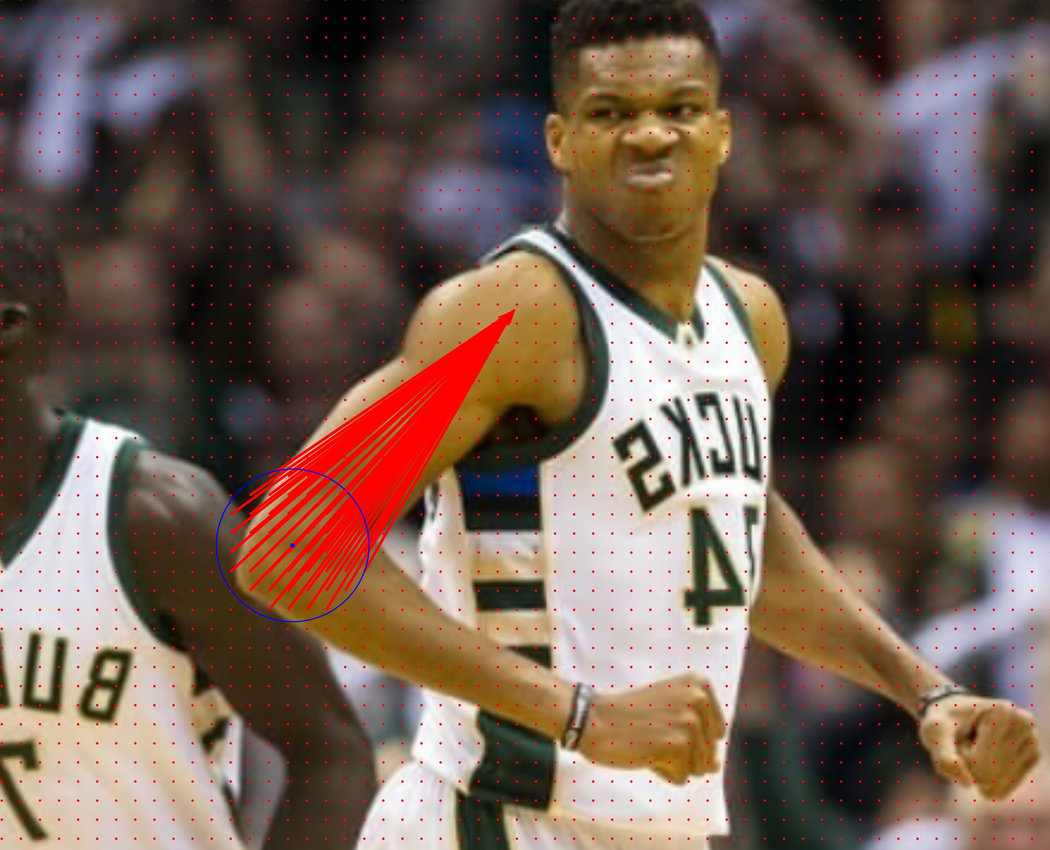} \\
    (a) & (b) & (c)\\
    \end{tabular}
    \vspace{-8pt}
    \caption{Mid-range offsets.
    (a) Initial mid-range offsets that starting around the \textsl{RightElbow} keypoint, they point towards the \textsl{RightShoulder} keypoint.
    (b) Mid-range offset refinement using the short-range offsets. 
    (c) Mid-range offsets after refinements.
    }
    \label{fig:offset_refinement}
    \vspace{-12pt}
\end{figure}

\paragraph{Mid-range pairwise offsets.} The local maxima in the score maps $h_k(x)$ serve as candidate positions for person keypoints, yet they carry no information about instance association. When multiple person instances are present in the image, we need a mechanism to ``connect the dots'' and group together the keypoints belonging to each individual instance. For this purpose, we add to our network a separate pairwise \emph{mid-range} 2-D offset field output $M_{k,l}(x)$ designed to connect pairs of keypoints. We compute $2(K-1)$ such offset fields, one for each directed edge connecting pairs $(k, l)$ of keypoints which are adjacent to each other in a tree-structured kinematic graph of the person, see Figs.~\ref{fig:pipeline} and~\ref{fig:offset_refinement}.
Specifically, the supervised training target for the pairwise offset field from the $k$-th to the $l$-th keypoint is given by $M_{k,l}(x) = (y_{j,l} - x) [x \in \mathcal{D}_R(y_{j,k})]$, since its purpose is to allow us to move from the $k$-th to the $l$-th keypoint of the same person instance $j$. During training, this target regression vector is only defined if both keypoints are present in the training example. We compute the average $L_1$ loss of the regression prediction errors over the source keypoint disks $x \in \mathcal{D}_R(y_{j,k})$ and back-propagate through the network.

\paragraph{Recurrent offset refinement.} Particularly for large person instances, the edges of the kinematic graph connect pairs of keypoints such as \textsl{RightElbow} and \textsl{RightShoulder} which may be several hundred pixels away in the image, making it hard to generate accurate regressions. We have successfully addressed this important issue by recurrently refining the mid-range pairwise offsets using the more accurate short-range offsets, specifically:
\begin{equation}
M_{k,l}(x) \leftarrow x' + S_{l}(x') \,, \mathrm{where} \,\, x' = M_{k,l}(x) \,,
\label{eq:offset_refinement}
\end{equation}
as illustrated in Fig.~\ref{fig:offset_refinement}. We repeat this refinement step twice in our experiments.
We employ bilinear interpolation to sample the short-range offset field at the intermediate position $x'$ and back-propagate the errors through it along both the mid-range and short-range input offset branches. We perform offset refinement at the resolution of CNN output activations (before upsamling to the original image resolution), making the process very fast. The offset refinement process drastically decreases the mid-range regression errors, as illustrated in Fig.\ref{fig:offset_refinement}.
This is a key novelty in our method, which greatly facilitates grouping and significantly improves results compared to previous papers \cite{cmu_mscoco, deeper_cut} which also employ pairwise displacements to associate keypoints.

\paragraph{Fast greedy decoding.} We have developed an extremely fast greedy decoding algorithm to group keypoints into detected person instances. We first create a priority queue, shared across all $K$ keypoint types, in which we insert the position $x_i$ and keypoint type $k$ of all local maxima in the Hough score maps $h_k(x)$ which have score above a threshold value (set to 0.01 in all reported experiments). These points serve as candidate seeds for starting a detection instance. We then pop elements out of the queue in descending score order. At each iteration, if the position $x_i$ of the current candidate detection seed of type $k$ is within a disk $\mathcal{D}_r(y_{j',k})$ of the corresponding keypoint of previously detected person instances $j'$, then we reject it; for this we use a non-maximum suppression radius of $r=10$ pixels. Otherwise, we start a new detection instance $j$ with the $k$-th keypoint at position $y_{j,k} = x_i$ serving as seed. We then follow the mid-range displacement vectors along the edges of the kinematic person graph to greedily connect pairs $(k, l)$ of adjacent keypoints, setting $y_{j,l} = y_{j,k} + M_{k,l}(y_{j,k})$.
\kevin{Do you look at the local score $p_k(x)$ at $y_{j,l}$ to check the prediction corresponds to a predicted point?}

It is worth noting that our decoding algorithm does not treat any keypoint type preferentially, in contrast to other techniques that always use the same keypoint type (\eg \textsl{Torso} or \textsl{Nose}) as seed for generating detections. Although we have empirically observed that the majority of detections in frontal facing person instances start from the more easily localizable facial keypoints, our approach can also handle robustly cases where a large portion of the person is occluded.
\kevin{Can you show an example of this?}

\subsubsection*{Keypoint- and instance-level detection scoring}

We have experimented with different methods to assign a keypoint- and instance-level score to the detections generated by our greedy decoding algorithm. Our first keypoint-level scoring method follows \cite{papandreou2017towards} and assigns to each keypoint a confidence score $s_{j,k} = h_k(y_{j,k})$. A drawback of this approach is that the well-localizable facial keypoints typically receive much higher scores than poorly localizable keypoints like the hip or knee. Our second approach attempts to calibrate the scores of the different keypoint types. It is motivated by the object keypoint similarity (OKS) evaluation metric used in the COCO keypoints task \cite{keypointchallenge}, which uses different accuracy thresholds $\kappa_k$ to penalize localization errors for different keypoint types.

Specifically, consider a detected person instance $j$ with keypoint coordinates $y_{j,k}$.
Let $\lambda_j$ be the square root of the area of the bounding box tightly containing all detected keypoints of the $j$-th person instance.
We define the \emph{Expected-OKS} score for the $k$-th keypoint by
\begin{equation}
s_{j,k} = E\{OKS_{j,k}\} = p_k(y_{j,k}) \int_{x \in \mathcal{D}_R(y_{j,k})} \hat{h}_k(x) \exp \left( -\frac{(x - y_{j,k})^2}{2 \lambda_j^2 \kappa_k^2} \right) dx \,,
\label{eq:expected_oks}
\end{equation}
where $\hat{h}_k(x)$ is the Hough score normalized in $\mathcal{D}_R(y_{j,k})$. The expected OKS keypoint-level score is the product of our confidence that the keypoint is present, times the OKS localization accuracy confidence, given the keypoint's presence.

We use the average of the keypoint scores as instance-level score $s_j^h = (1/K) \sum_k s_{j,k}$, followed by non-maximum suppression (NMS).
\kevin{Why $s_j^h$ and not $s_j$?}
We have experimented both with hard OKS-based NMS \cite{papandreou2017towards} as well as a soft-NMS scheme adapted for the keypoints tasks from \cite{bodla2017soft}, where we use as final instance-level score the sum of the scores of the keypoints that have not already been claimed by higher scoring instances, normalized by the total number of keypoints: 
\begin{equation}
s_j = (1/K) \sum_{k=1:K} s_{j,k} [\norm{y_{j,k} - y_{j',k}} > r , \textrm{for every} \,\, j' < j] \,,
\label{eq:instance_score_soft_nms}
\end{equation}
where $r=10$ is the NMS-radius. In the experiments reported in Sec.~\ref{sec:experiments} we report results with the best performing Expected-OKS scoring and soft-NMS but we include an ablation study in Appendix~\ref{sec:ablations}.

\subsection{Instance-level person segmentation}

\begin{figure}[t]
    \centering
    \begin{tabular}{cccc}
    \includegraphics[width=0.23\linewidth]{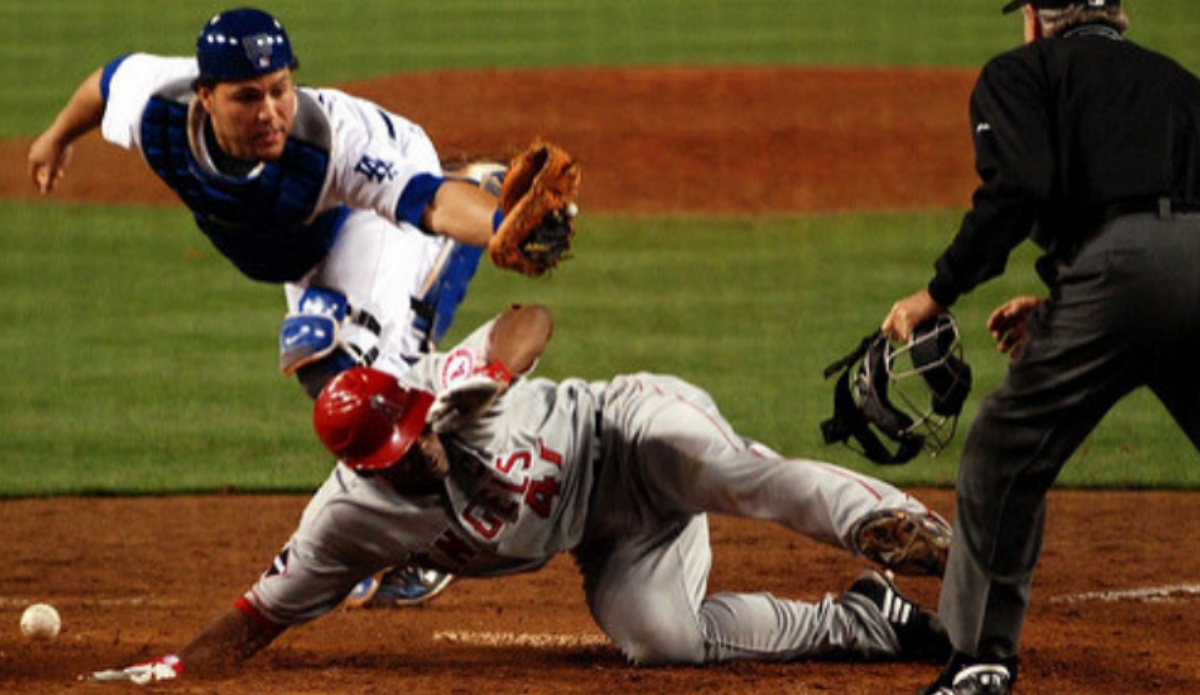} &
    \includegraphics[width=0.23\linewidth]{figs/example/segments.jpg} &
    \includegraphics[width=0.23\linewidth]{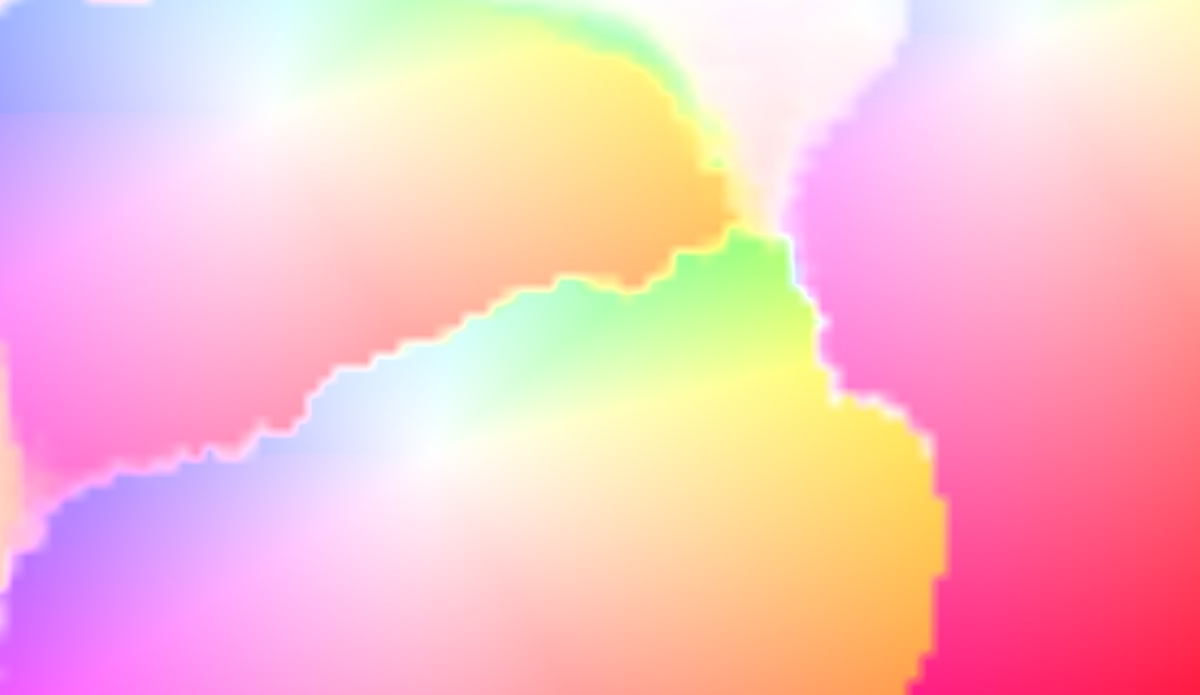} &
    \includegraphics[width=0.23\linewidth]{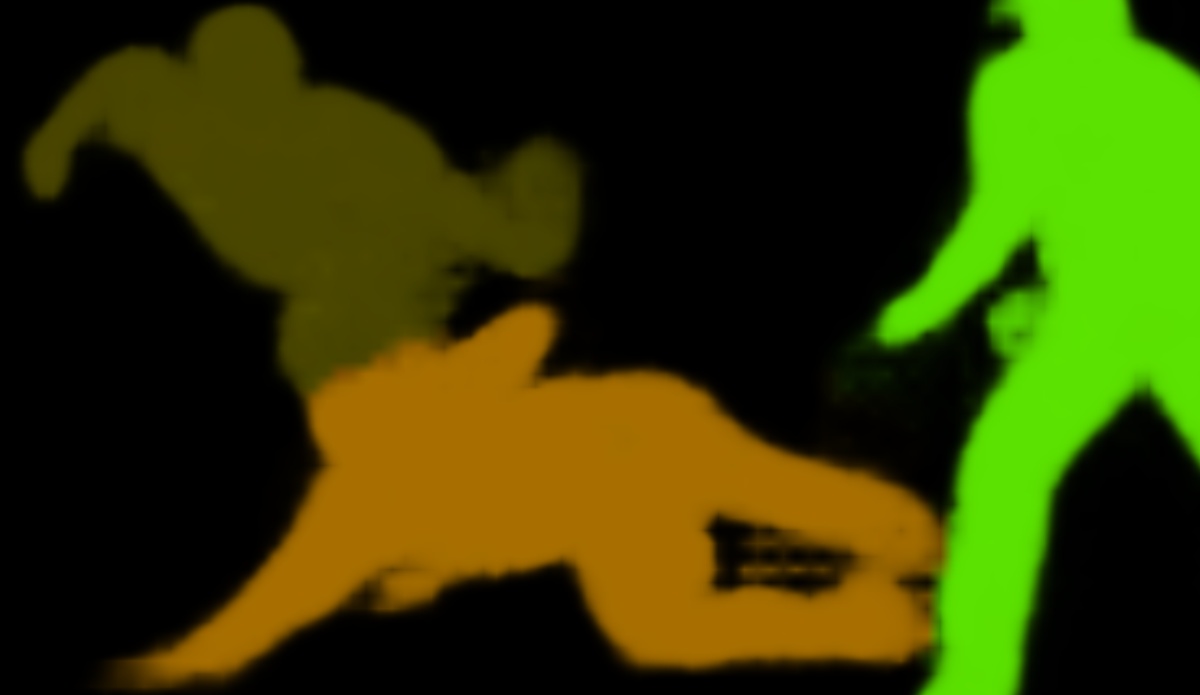}\\
    (a) & (b) & (c) & (d)\\
    \end{tabular}
    \caption{From semantic to instance segmentation: (a) Image; (b) person segmentation; (c) basins of attraction defined by the long-range offsets to the \textsl{Nose} keypoint; (d) instance segmentation masks.
    \kevin{Explain}
    }
    \label{fig:embedding_instance}
\end{figure}

Given the set of keypoint-level person instance detections, the task of the instance segmentation stage is to identify pixels that belong to people (recognition) and associate them with the detected person instances (grouping). We describe next the respective semantic segmentation and association modules, illustrated in Fig.~\ref{fig:embedding_instance}.

\begin{figure}[t]
    \centering
    \begin{tabular}{ccc}
    \includegraphics[width=0.30\linewidth]{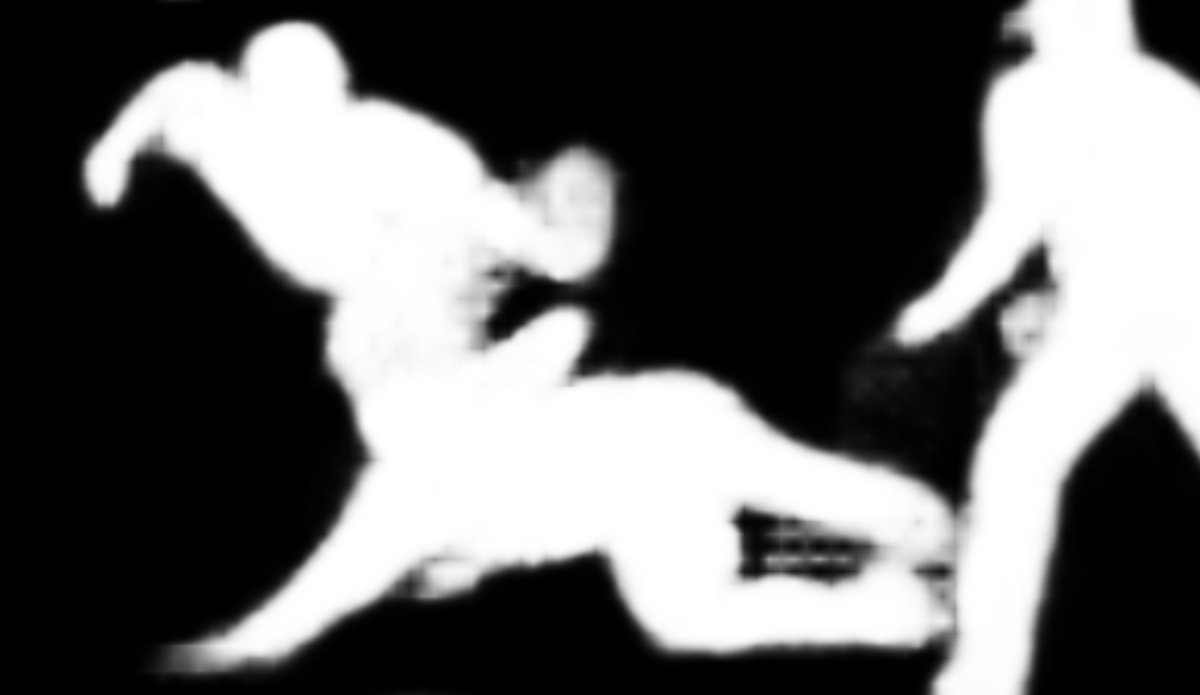} &
    \includegraphics[width=0.30\linewidth]{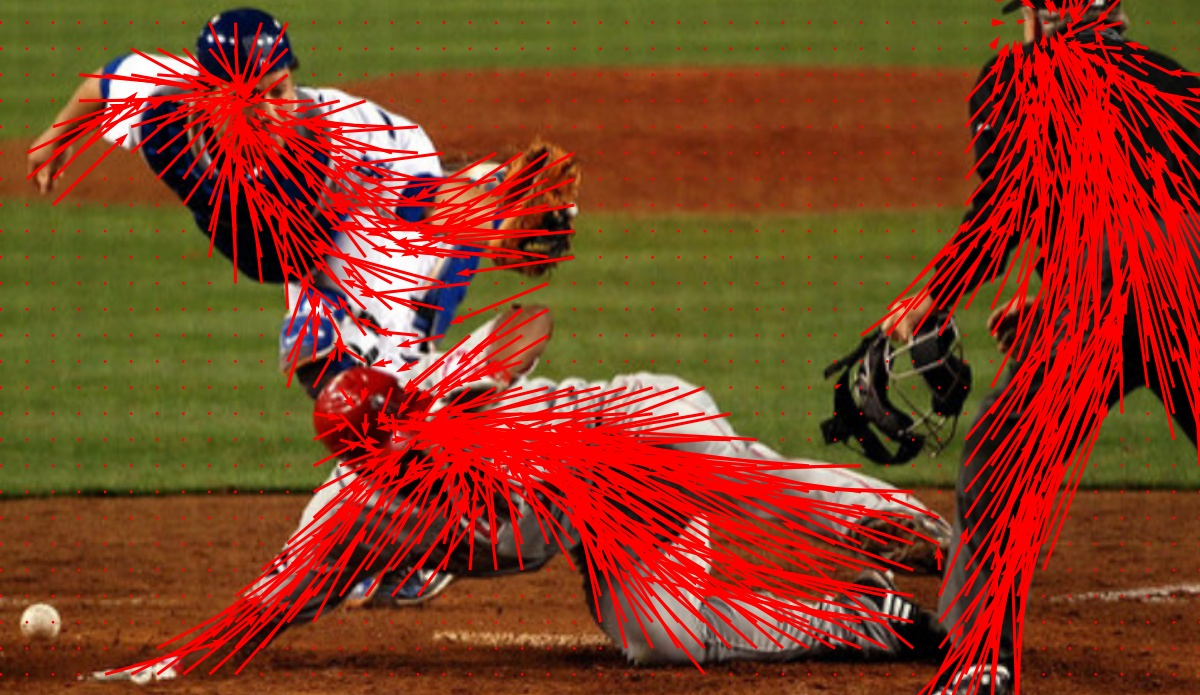} &
    \includegraphics[width=0.30\linewidth]{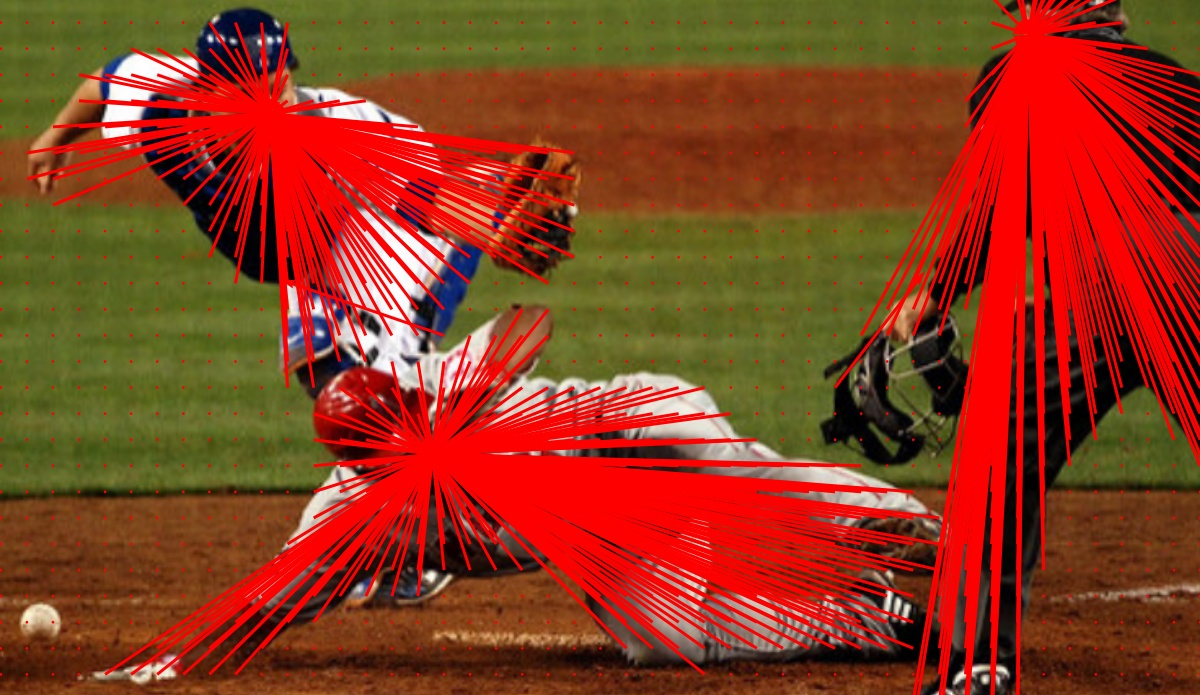}\\
    (a) & (b) & (c)\\
    \end{tabular}
    \vspace{-8pt}
    \caption{Long-range offsets defined in the person segmentation mask.
    (a) Estimated person segmentation map.
    (b) Initial long range offsets for the \textsl{Nose} destination keypoint: each pixel in the foreground of the person segmentation mask points towards the \textsl{Nose} keypoint of the instance that it belongs to.
    \kevin{Can you use different colors for the 3 different instances?}
    (c) Long-range offsets after their refinements with the short-range offsets.}
    \label{fig:embedding_refinement}
\end{figure}

\subsubsection*{Semantic person segmentation}

We treat semantic person segmentation in the standard fully-convolutional fashion \cite{long2014fully, chen2016deeplab}. We use a simple semantic segmentation head consisting of a single 1x1 convolutional layer that performs dense logistic regression and compute at each image pixel $x_i$ the probability $p_S(x_i)$ that it belongs to at least one person. During training, we compute and backpropagate the average of the logistic loss over all image regions that have been annotated with person segmentation maps (in the case of COCO we exclude the crowd person areas).

\subsubsection*{Associating segments with instances via geometric embeddings}

The task of this module is to associate each person pixel identified by the semantic segmentation module with the keypoint-level detections produced by the person detection and pose estimation module.

Similar to \cite{newell2017associative, fathi2017semantic, de2017semantic}, we follow the embedding-based approach for this task. In this framework, one computes an embedding vector $G(x)$ at each pixel location, followed by clustering to obtain the final object instances. In previous works, the representation is typically learned by computing pairs of embedding vectors at different image positions and using a loss function designed to attract the two embedding vectors if they both come from the same object instance and repel them if they come from different person instances. This typically leads to embedding representations which are difficult to interpret and involves solving a hard learning problem which requires careful selection of the loss function and tuning several hyper-parameters such as the pair sampling protocol.

Here, we opt instead for a considerably simpler, geometric approach. At each image position $x$ inside the segmentation mask of an annotated person instance $j$ with 2-D keypoint positions $y_{j,k}, k=1,\dots,K$, we define the \emph{long-range offset} vector $L_k(x) = y_{j,k} - x$ which points from the image position $x$ to the position of the $k$-th keypoint of the corresponding instance $j$.
(This is very similar to the short-range prediction task, except the dynamic range is different, since we require the network to predict from any pixel inside the person, not just from inside a disk near the keypoint.
Thus these are like two "specialist" networks. Performance is worse when we use the same network
for both kinds of tasks. \kevin{Check this!})
We compute $K$ such 2-D vector fields, one for each keypoint type. During training, we penalize the long-range offset regression errors using the $L_1$ loss, averaging and back-propagating the errors only at image positions $x$ which belong to a single person object instance. We ignore background areas, crowd regions, and pixels which are covered by two or more person masks.

The long-range prediction task is challenging, especially for large object instances that may cover the whole image.
As in Sec.~\ref{sec:keypoint_grouping}, we recurrently refine the long-range offsets, twice by themselves and then twice by the short-range offsets
\begin{equation}
L_k(x) \leftarrow x' + L_k(x') \,, x' = L_k(x) \,\,\,
\mathrm{and} \,\,\,
L_k(x) \leftarrow x' + S_k(x') \,, x' = L_k(x) \,,
\label{eq:embedding_refinement}
\end{equation}
back-propagating through the bilinear warping function during training. Similarly with the mid-range offset refinement in Eq.~\ref{eq:offset_refinement}, recurrent long-range offset refinement dramatically improves the long-range offset prediction accuracy.

In Fig.~\ref{fig:embedding_refinement} we illustrate the long-range offsets corresponding to the \textsl{Nose} keypoint as computed by our trained CNN for an example image. We see that the long-range vector field effectively partitions the image plane into basins of attraction for each person instance. This motivates us to define as embedding representation for our instance association task the $2 \cdot K$ dimensional vector $G(x) = (G_k(x))_{k=1, \dots, K}$ with components $G_k(x)= x + L_k(x)$.

Our proposed embedding vector has a very simple geometric interpretation: At each image position $x_i$ semantically recognized as a person instance, the embedding $G(x_i)$ represents our local estimate for the absolute position of every keypoint of the person instance it belongs to,
i.e., it represents the predicted shape of the person.
This naturally suggests shape metric as candidates for computing distances in our proposed embedding space. In particular, in order to decide if the person pixel $x_i$ belongs to the $j$-th person instance, we compute the embedding distance metric
\begin{equation}
    D_{i,j} = \frac{1}{\sum_k p_k(y_{j,k})} \sum_{k=1}^K  p_k(y_{j,k}) \frac{1}{\lambda_j}\norm{G_k(x_i) - y_{j,k}} \,,
\label{eq:distance_metric}
\end{equation}
where $y_{j,k}$ is the position of the $k$-th detected keypoint in the $j$-th instance and $p_k(y_{j,k})$ is the probability that it is present. Weighing the errors by the keypoint presence probability allows us to discount discrepancies in the two shapes due to missing keypoints. Normalizing the errors by the detected instance scale $\lambda_j$ allows us to compute a scale invariant metric. We set $\lambda_j$ equal to the square root of the area of the bounding box tightly containing all detected keypoints of the $j$-th person instance. We emphasize that because we only need to compute the distance metric between the $N_S$ pixels and the $M$ person instances, our algorithm is very fast in practice, having complexity $\mathcal{O}(N_S * M)$ instead of $\mathcal{O}(N_S * N_S)$ of standard embedding-based segmentation techniques which, at least in principle, require computation of embedding vector distances for all pixel pairs.

To produce the final instance segmentation result: (1) We find all positions $x_i$ marked as person in the semantic segmentation map, \ie those pixels that have semantic segmentation probability $p_S(x_i) \ge 0.5$. (2) We associate each person pixel $x_i$ with every detected person instance $j$ for which the embedding distance metric satisfies $D_{i,j} \le t$; we set the relative distance threshold $t=0.25$ for all reported experiments. It is important to note that the pixel-instance assignment is non-exclusive: Each person pixel may be associated with more than one detected person instance (which is particularly important when doing soft-NMS in the detection stage) or it may remain an orphan (\eg, a small false positive region produced by the segmentation module). We use the same instance-level score produced by the previous person detection and pose estimation stage to also evaluate on the COCO segmentation task and obtain average precision performance numbers.


\subsection{Imputing missing keypoint annotations}
\label{sec:keypoint_imputation}

The standard COCO dataset does not contain keypoint annotations in the training set for the small person instances, and ignores them during model evaluation. However, it contains segmentation annotations and evaluates mask predictions for those small instances. Since training our geometric embeddings requires keypoint annotations for training, we have run the single-person pose estimator of \cite{papandreou2017towards} (trained on COCO data alone) in the COCO training set on image crops around the ground truth box annotations of those small person instances to impute those missing keypoint annotations. We treat those imputed keypoints as regular training annotations during our PersonLab model training. Naturally, this missing keypoint imputation step is particularly important for our COCO instance segmentation performance on small person instances, as shown in Appendix~\ref{sec:ablations}. We emphasize that, unlike \cite{radosavovic2017data}, we do not use any data beyond the COCO \emph{train} split images and annotations in this process. Data distillation on additional images as described in \cite{radosavovic2017data} may yield further improvements.


\kevin{Fix me}



\section{Experimental evaluation}
\label{sec:experiments}

\newlength{\viswidthA}
\setlength{\viswidthA}{2.8cm}
\newlength{\viswidthB}
\setlength{\viswidthB}{1.85cm}
\newlength{\viswidthC}
\setlength{\viswidthC}{2.2cm}
\newlength{\visheightA}
\setlength{\visheightA}{3.55cm}

\subsection{Experimental Setup}
\label{sec:experimental_setup}

\paragraph{Dataset and Tasks}

We evaluate the proposed PersonLab system on the standard COCO keypoints task~\cite{keypointchallenge} and on COCO instance segmentation~\cite{lin2014microsoft} for the person class alone. For all reported results we only use COCO data for model training (in addition to  Imagenet pretraining). Our \emph{train} set is the subset of the 2017 COCO training set images that contain people (64115 images). Our \emph{val} set coincides with the 2017 COCO validation set (5000 images). We only use \emph{train} for training and evaluate on either \emph{val} or the \emph{test-dev} split (20288 images).

\paragraph{Model training details}

We report experimental results with models that use either ResNet-101 or ResNet-152 CNN backbones \cite{He2016ResNets} pretrained on the Imagenet classification task~\cite{imagenet2015}. We discard the last Imagenet classification layer and add 1x1 convolutional layers for each of our model-specific layers. During model training, we randomly resize a square box tightly containing the full image by a uniform random scale factor between 0.5 and 1.5, randomly translate it along the horizontal and vertical directions, and left-right flip it with probability 0.5. We sample and resize the image crop contained under the resulting perturbed box to an 801x801 image that we feed into the network. We use a batch size of 8 images distributed across 8 Nvidia Tesla P100 GPUs in a single machine and perform synchronous training for 1M steps with stochastic gradient descent with constant learning rate equal to 1e-3, momentum value set to 0.9, and Polyak-Ruppert model parameter averaging. We employ batch normalization \cite{ioffe2015batch} but fix the statistics of the ResNet activations to their Imagenet values. Our ResNet CNN network backbones have nominal output stride (\ie, ratio of the input image to output activations size) equal to 32 but we reduce it to 16 during training and 8 during evaluation using atrous convolution \cite{chen2016deeplab}. During training we also make model predictions using as features activations from a layer in the middle of the network, which we have empirically observed to accelerate training. To balance the different loss terms we use weights equal to $(4, 2, 1, 1/4, 1/8)$ for the heatmap, segmentation, short-range, mid-range, and long-range offset losses in our model. For evaluation we report both single-scale results (image resized to have larger side 1401 pixels) and multi-scale results (pyramid with images having larger side 601, 1201, 1801, 2401 pixels). We have implemented our system in Tensorflow \cite{tensorflow2015-whitepaper}. All reported numbers have been obtained with a single model without ensembling.

\subsection{COCO person keypoints evaluation}
\label{sec:keypoint_evaluation} 

Table~\ref{table:coco_keypoint_results_testdev} shows our system's person keypoints performance on COCO \emph{test-dev}. Our single-scale inference result is already better than the results of the CMU-Pose~\cite{cmu_mscoco} and Associative Embedding~\cite{newell2017associative} bottom-up methods, even when they perform multi-scale inference and refine their results with a single-person pose estimation system applied on top of their bottom-up detection proposals. Our results also outperform top-down methods like Mask-RCNN~\cite{he2017mask} and G-RMI~\cite{papandreou2017towards}. Our best result with 0.687 AP is attained with a ResNet-152 based model and multi-scale inference. Our result is still behind the winners of the 2017 keypoints challenge (Megvii) \cite{chen2017cascaded} with 0.730 AP,
but they used a carefully tuned two-stage, top-down model that also builds on a significantly more powerful CNN backbone.

\begin{table*}[t]
\centering
\caption{Performance on the COCO keypoints \textbf{test-dev} split.}
\label{table:coco_keypoint_results_testdev}
\scalebox{0.7}{
\begin{tabular}{l|ccccc|ccccc}
    &$AP$ & $AP^{.50}$ & $AP^{.75}$ & $AP^M$  & $AP^L$ & $AR$ & $AR^{.50}$ & $AR^{.75}$  & $AR^M$ & $AR^L$ \\
  \hline
Bottom-up methods: \\ 
CMU-Pose~\cite{cmu_mscoco} (+refine) & 0.618 & 0.849 & 0.675 & 0.571 & 0.682 & 0.665 & 0.872 & 0.718 & 0.606 & 0.746 \\
Assoc. Embed.~\cite{newell2017associative} (multi-scale) & 0.630 & 0.857 & 0.689 & 0.580 & 0.704 & - & - & - & - & - \\
Assoc. Embed.~\cite{newell2017associative}  (mscale, refine) & 0.655 & 0.879 & 0.777 & 0.690 & 0.752 & 0.758 & 0.912 & 0.819 & 0.714 & 0.820 \\ \hline
Top-down methods: \\ 
Mask-RCNN~\cite{he2017mask} & 0.631 & 0.873 & 0.687 & 0.578 & 0.714 & 0.697 & 0.916 & 0.749 & 0.637 & 0.778 \\ 
G-RMI \emph{COCO-only} ~\cite{papandreou2017towards}  & 0.649 & 0.855 & 0.713 & 0.623 & 0.700 & 0.697 & 0.887 & 0.755 & 0.644 & 0.771 \\ \hline\hline
PersonLab (ours): \\ 
ResNet101 (single-scale) &  0.655 & 0.871 & 0.714 & 0.613 & 0.715 & 0.701 & 0.897 & 0.757 & 0.650 & 0.771 \\
ResNet152  (single-scale) &  \textbf{0.665} & 0.880 & 0.726 & 0.624 & 0.723 & 0.710 & 0.903 & 0.766 & 0.661 & 0.777 \\ \hline
ResNet101  (multi-scale) &  0.678 & 0.886 & 0.744 & 0.630 & 0.748 & 0.745 & 0.922 & 0.804 & 0.686 & 0.825 \\
ResNet152 (multi-scale) &  \textbf{0.687} & 0.890 & 0.754 & 0.641 & 0.755 & 0.754 & 0.927 & 0.812 & 0.697 &0.830 \\
\end{tabular}}
\end{table*}

\subsection{COCO person instance segmentation evaluation}
\label{sec:segmentation_evaluation} 

Tables~\ref{table:coco_segmentation_results_testdev} and ~\ref{table:coco_segmentation_results_val} show our person instance segmentation results on COCO \emph{test-dev} and \emph{val}, respectively. We use the small-instance missing keypoint imputation technique of Sec.~\ref{sec:keypoint_imputation} for the reported instance segmentation experiments, which significantly increases our performance for small objects. Our results without missing keypoint imputation are shown in Appendix~\ref{sec:ablations}.

Our method only produces segmentation results for the person class, since our system is keypoint-based and thus cannot be applied to the other COCO classes. The standard COCO instance segmentation evaluation allows for a maximum of 100 proposals per image for all 80 COCO classes. For a fair comparison when comparing with previous works, we report \emph{test-dev} results of our method with a maximum of 20 person proposals per image, which is the convention also adopted in the standard COCO person keypoints evaluation protocol. For reference, we also report the \emph{val} results of our best model when allowed to produce 100 proposals.
 
We compare our system with the person category results of top-down instance segmentation methods. As shown in Table~\ref{table:coco_segmentation_results_testdev}, our method on the test split
outperforms FCIS~\cite{dai2017fully} in both single-scale and multi-scale inference settings.
As shown in Table~\ref{table:coco_segmentation_results_val}, our performance on the val split is similar to that of Mask-RCNN~\cite{he2017mask} on medium and large person instances, but worse on small person instances.
However, we emphasize that our method is the first box-free, bottom-up instance segmentation method to report experiments on the COCO instance segmentation task.

\begin{table*}[t]
\centering
\caption{Performance on COCO Segmentation (Person category) \textbf{test-dev} split. Our person-only results have been obtained with 20 proposals per image. The person category FCIS eval results have been communicated by the authors of~\cite{dai2017fully}.}
\label{table:coco_segmentation_results_testdev}
\scalebox{0.7}{
\begin{tabular}{l|cccccc|cccccc}
  & $AP$ & $AP^{50}$ & $AP^{75}$ & $AP^S$ & $AP^M$ & $AP^L$ & $AR^1$ & $AR^{10}$ & $AR^{100}$ & $AR^{S}$ & $AR^{M}$ & $AR^L$ \\
 \hline
FCIS (baseline) ~\cite{dai2017fully}  & 0.334 & 0.641 & 0.318 & 0.090 & 0.411 & 0.618 & 0.153 & 0.372 & 0.393 & 0.139 & 0.492 & 0.688 \\
FCIS (multi-scale) ~\cite{dai2017fully}  & 0.386 & 0.693 & 0.410 & 0.164 & 0.481 & 0.621 & 0.161 & 0.421 & 0.451 & 0.221 & 0.562 & 0.690 \\ \hline\hline
PersonLab (ours):  \\ 
ResNet101 (1-scale, 20 prop) &  0.377 & 0.659 & 0.394 & 0.166 & 0.480 & 0.595 & 0.162 & 0.415 & 0.437 & 0.207 & 0.536 & 0.690 \\
ResNet152 (1-scale, 20 prop) &  \textbf{0.385} & 0.668 & 0.404 & 0.172 & 0.488 & 0.602 & 0.164 & 0.422 & 0.444 & 0.215 & 0.544 & 0.698 \\ \hline
ResNet101 (mscale, 20 prop) & 0.411 & 0.686 & 0.445 & 0.215 & 0.496 & 0.626 & 0.169 & 0.453 & 0.489 & 0.278 & 0.571 & 0.735  \\
ResNet152 (mscale, 20 prop) & \textbf{0.417} & 0.691 & 0.453 & 0.223 & 0.502 & 0.630 & 0.171 & 0.461 & 0.497 & 0.287 & 0.578 & 0.742  \\
\end{tabular}}
\end{table*}

\begin{table*}[t]
\centering
\caption{Performance on COCO Segmentation (Person category) \textbf{val} split. The Mask-RCNN ~\cite{he2017mask} person results have been produced by the ResNet-101-FPN version of their publicly shared model (which achieves 0.359 AP across all COCO classes).}
\label{table:coco_segmentation_results_val}
\scalebox{0.7}{
\begin{tabular}{l|cccccc|cccccc}
                  & $AP$    & $AP^{50}$ & $AP^{75}$ & $AP^S$ & $AP^M$ & $AP^L$ & $AR^1$ & $AR^{10}$ & $AR^{100}$ & $AR^{S}$ & $AR^{M}$ & $AR^L$ \\ \hline
Mask-RCNN~\cite{he2017mask} & 0.455 & 0.798 & 0.472 & 0.239 & 0.511 & 0.611 & 0.169 & 0.477 & 0.530 & 0.350 & 0.596 & 0.721 \\ \hline\hline
PersonLab (ours):  \\ 
ResNet101 (1-scale, 20 prop)    & 0.382 & 0.661 & 0.397 & 0.164 & 0.476 & 0.592 & 0.162 & 0.416 & 0.439 & 0.204 & 0.532 & 0.681  \\
ResNet152 (1-scale, 20 prop)    & 0.387 & 0.667 & 0.406 & 0.169 & 0.483 & 0.595 & 0.163 & 0.423 & 0.446 & 0.213 & 0.539 & 0.686  \\ \hline
ResNet101 (mscale, 20 prop) &  0.414 & 0.684 & 0.447 & 0.213 & 0.492 & 0.621 & 0.170 & 0.454 & 0.492 & 0.278 & 0.566 & 0.728 \\
ResNet152 (mscale, 20 prop) &  0.418 & 0.688 & 0.455 & 0.219 & 0.497 & 0.621 & 0.170 & 0.460 & 0.497 & 0.284 & 0.573 & 0.730 \\ \hline
ResNet152 (mscale, 100 prop) & 0.429 & 0.711 & 0.467 & 0.235 & 0.511 & 0.623 & 0.170 & 0.460 & 0.539 & 0.346 & 0.612 & 0.741  \\
\end{tabular}}
\end{table*}

\subsubsection*{Qualitative evaluation}
\label{sec:qualitative_evaluation}

\begin{figure}
    \begin{center}
        \centering
    \adjustbox{width=\viswidthA}{\includegraphics{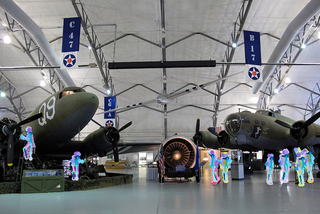}}
    \adjustbox{width=\viswidthA}{\includegraphics{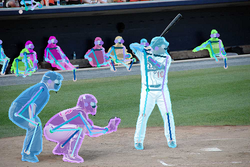}}
    \adjustbox{width=\viswidthA}{\includegraphics{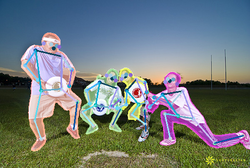}}
    \adjustbox{width=\viswidthA}{\includegraphics{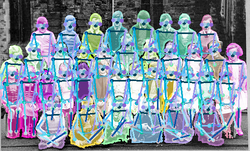}}
    
    \vspace{1mm}
    \adjustbox{width=\viswidthA}{\includegraphics{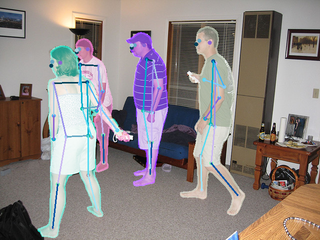}}
    \adjustbox{width=\viswidthA}{\includegraphics{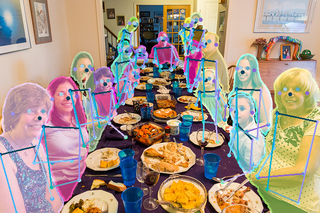}}
    \adjustbox{width=\viswidthA}{\includegraphics{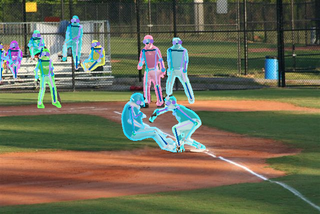}}
    \adjustbox{width=\viswidthA}{\includegraphics{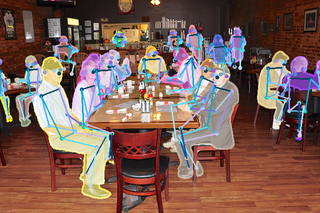}}

    \vspace{1mm}
    \adjustbox{width=\viswidthA}{\includegraphics{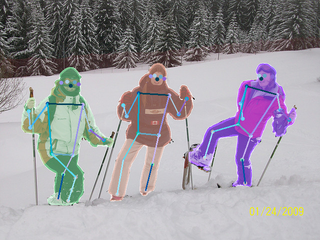}}
    \adjustbox{width=\viswidthA}{\includegraphics{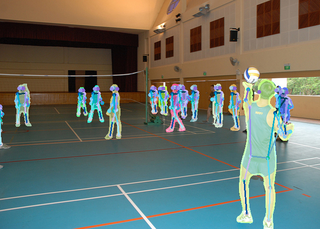}}
    \adjustbox{width=\viswidthA}{\includegraphics{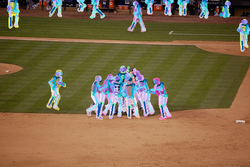}}
    \adjustbox{width=\viswidthA}{\includegraphics{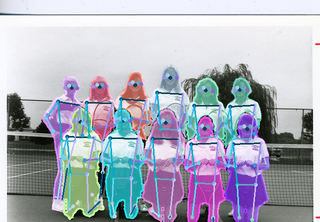}}
    
    \vspace{1mm}
    \adjustbox{width=\viswidthA}{\includegraphics{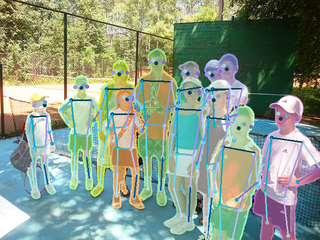}}
    \adjustbox{width=\viswidthA}{\includegraphics{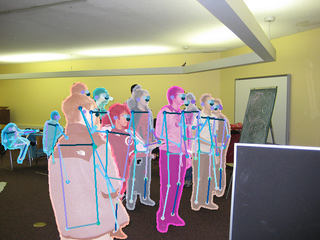}}
    \adjustbox{width=\viswidthA}{\includegraphics{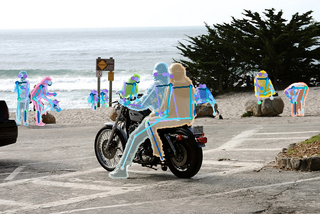}}
    \adjustbox{width=\viswidthA}{\includegraphics{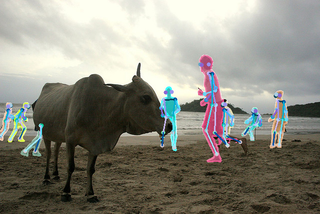}}
    
    \vspace{1mm}
    \adjustbox{height=\visheightA}{\includegraphics{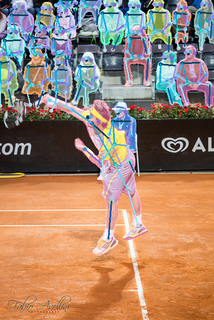}}
    \adjustbox{height=\visheightA}{\includegraphics{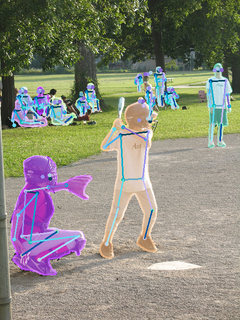}}
    \adjustbox{height=\visheightA}{\includegraphics{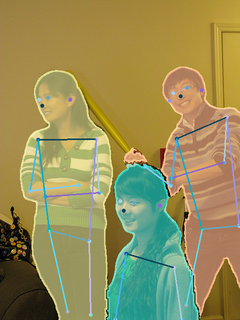}}
    \adjustbox{height=\visheightA}{\includegraphics{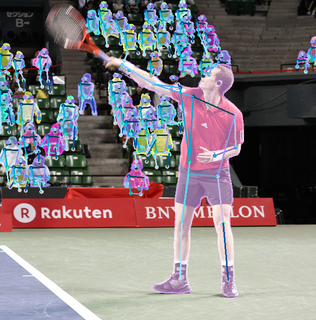}}
    
    \vspace{1mm}
    \adjustbox{height=\viswidthC}{\includegraphics{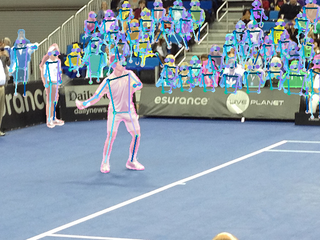}}
    \adjustbox{height=\viswidthC}{\includegraphics{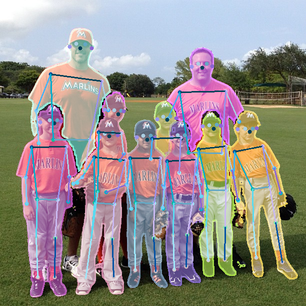}}
    \adjustbox{height=\viswidthB}{\includegraphics{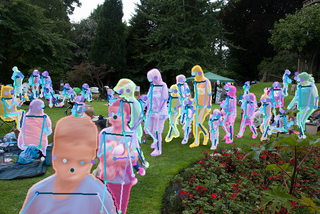}}
    \adjustbox{height=\viswidthB}{\includegraphics{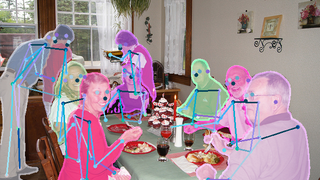}}

    \vspace{1mm}
    \adjustbox{width=\viswidthA}{\includegraphics{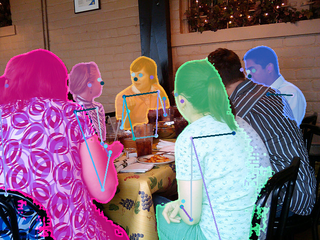}}
    \adjustbox{width=\viswidthA}{\includegraphics{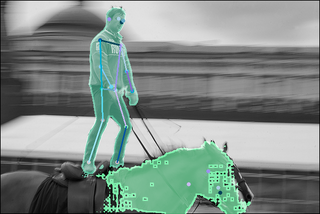}}
    \adjustbox{width=\viswidthA}{\includegraphics{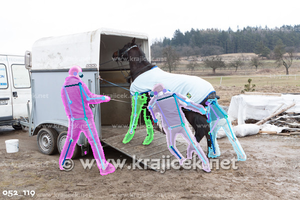}}
    \adjustbox{width=\viswidthA}{\includegraphics{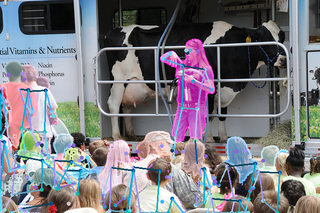}}
    
    \end{center}    
  \caption{Visualization on COCO \emph{val} images. The last row shows some failure cases: missed key point detection, false positive key point detection, and missed segmentation.}
  \label{fig:vis_results}
\end{figure}

In Fig.~\ref{fig:vis_results} we show representative person pose and instance segmentation results on COCO \emph{val} images produced by our model with single-scale inference.

\section{Conclusions}

We have developed a bottom-up model which jointly addresses the problems of person detection, pose estimation, and instance segmentation using a unified part-based modeling approach. We have demonstrated the effectiveness of the proposed method on the challenging COCO person keypoint and instance segmentation tasks. A key limitation of the proposed method is its reliance on keypoint-level annotations for training on the instance segmentation task. In the future, we plan to explore ways to overcome this limitation, via weakly supervised part discovery.

\appendix

\section{Ablation Experiments}
\label{sec:ablations}

We perform a series of ablation experiments examining the effect of different model choices to the system's performance. In all corresponding Tables we indicate with boldface type the model variant employed in the results reported in Sec.~\ref{sec:experiments}. For all ablation experiments we use a ResNet-101 model and single-scale inference.

\subsection{Ablation: Input image size and activation output stride}

Given a trained PersonLab model, we have two key knobs that we can use to control its speed/accuracy tradeoff. Table~\ref{table:coco_keypoint_results_val_input_size_output_stride} shows our system's 
person keypoints performance on COCO \emph{val} when varying the input image size (we resize the input image so that its largest side equals the specified value) and the output activation stride (we control the output stride by employing atrous convolution; larger output stride value leads to faster inference and smaller output stride value improves the accuracy of the results).

We observe that model performance increases significantly when we compute output activations more densely, using atrous convolution to decrease the output stride from 32 down to 16 pixels. Decreasing the output stride further from 16 down to 8 pixels brings a further small performance improvement, yet it significantly increases the model's computation cost. For large person instances, we get reasonably good keypoint AP performance for as small as 601 or 801 pixels input image size. However, accurately capturing small person instances requires us to use higher resolution input images.

In terms of model inference speed as measured on a Titan X using as input a 801x529 image, inference time is 341 msec for output stride equal to 32, 355 msec for output stride equal to 16, and 464 msec for output stride equal to 8. This refers to end-to-end timing to produce both the keypoint and instance segmentation final outputs. We see that using output stride equal to 16 pixels strikes an excellent speed-accuracy tradeoff.

\begin{table*}[t]
\centering
\caption{PersonLab performance on the COCO keypoints \textbf{val} split. Single-scale ResNet-101 model evaluation for varying input image size and activation output stride.}
\label{table:coco_keypoint_results_val_input_size_output_stride}
\scalebox{0.8}{
\begin{tabular}{l|ccccc|ccccc}
        &$AP$ & $AP^{.50}$ & $AP^{.75}$ & $AP^M$  & $AP^L$ & $AR$ & $AR^{.50}$ & $AR^{.75}$  & $AR^M$ & $AR^L$ \\
\hline
\multicolumn{2}{l}{Output stride 32:} \\
Input  401 & 0.356 & 0.553 & 0.358 & 0.157 & 0.625 & 0.384 & 0.572 & 0.389 & 0.176 & 0.670 \\
Input  601 & 0.481 & 0.700 & 0.500 & 0.310 & 0.712 & 0.516 & 0.723 & 0.536 & 0.344 & 0.752 \\
Input  801 & 0.559 & 0.780 & 0.595 & 0.433 & 0.736 & 0.598 & 0.807 & 0.633 & 0.470 & 0.777 \\
Input 1001 & 0.609 & 0.830 & 0.655 & 0.519 & 0.740 & 0.649 & 0.851 & 0.693 & 0.556 & 0.780 \\
Input 1201 & 0.630 & 0.842 & 0.684 & 0.565 & 0.731 & 0.673 & 0.867 & 0.723 & 0.602 & 0.774 \\
Input 1401 & 0.641 & 0.850 & 0.694 & 0.591 & 0.720 & 0.684 & 0.871 & 0.733 & 0.628 & 0.765 \\
Input 1601 & 0.639 & 0.849 & 0.696 & 0.603 & 0.703 & 0.685 & 0.874 & 0.738 & 0.639 & 0.751 \\
Input 1801 & 0.634 & 0.840 & 0.690 & 0.609 & 0.681 & 0.682 & 0.868 & 0.734 & 0.645 & 0.736 \\
\hline
\multicolumn{2}{l}{Output stride 16:} \\
Input  401 & 0.400 & 0.603 & 0.413 & 0.206 & 0.662 & 0.432 & 0.622 & 0.448 & 0.229 & 0.710 \\
Input  601 & 0.532 & 0.760 & 0.563 & 0.386 & 0.731 & 0.570 & 0.784 & 0.602 & 0.423 & 0.775 \\
Input  801 & 0.600 & 0.821 & 0.643 & 0.497 & 0.746 & 0.641 & 0.846 & 0.683 & 0.535 & 0.789 \\
Input 1001 & 0.636 & 0.850 & 0.688 & 0.559 & 0.750 & 0.677 & 0.873 & 0.727 & 0.595 & 0.793 \\
Input 1201 & 0.651 & 0.860 & 0.705 & 0.593 & 0.740 & 0.695 & 0.884 & 0.746 & 0.630 & 0.786 \\
Input 1401 & 0.656 & 0.859 & 0.714 & 0.611 & 0.728 & 0.701 & 0.885 & 0.754 & 0.647 & 0.779 \\
Input 1601 & 0.654 & 0.858 & 0.714 & 0.622 & 0.708 & 0.701 & 0.885 & 0.756 & 0.659 & 0.762 \\
Input 1801 & 0.645 & 0.847 & 0.702 & 0.624 & 0.686 & 0.696 & 0.878 & 0.750 & 0.660 & 0.746 \\
\hline
\multicolumn{2}{l}{Output stride 8:} \\
Input  401 & 0.405 & 0.599 & 0.425 & 0.220 & 0.667 & 0.433 & 0.613 & 0.452 & 0.232 & 0.709 \\
Input  601 & 0.541 & 0.764 & 0.577 & 0.406 & 0.733 & 0.577 & 0.787 & 0.613 & 0.435 & 0.774 \\
Input  801 & 0.612 & 0.824 & 0.658 & 0.517 & 0.752 & 0.650 & 0.849 & 0.693 & 0.550 & 0.790 \\
Input 1001 & 0.646 & 0.854 & 0.698 & 0.576 & 0.753 & 0.684 & 0.873 & 0.735 & 0.608 & 0.793 \\
Input 1201 & 0.659 & 0.862 & 0.711 & 0.607 & 0.743 & 0.700 & 0.885 & 0.750 & 0.639 & 0.786 \\
\textbf{Input 1401} & 0.665 & 0.862 & 0.719 & 0.623 & 0.732 & 0.707 & 0.887 & 0.757 & 0.656 & 0.779 \\
Input 1601 & 0.662 & 0.861 & 0.718 & 0.632 & 0.712 & 0.706 & 0.885 & 0.755 & 0.665 & 0.765 \\
Input 1801 & 0.652 & 0.855 & 0.714 & 0.634 & 0.690 & 0.701 & 0.881 & 0.755 & 0.667 & 0.749 \\
\end{tabular}}
\end{table*}

\subsection{Ablation: Keypoint scoring and non-maximum suppression}

We examine the effect of the two keypoint scoring mechanisms examined in Sec.~\ref{sec:keypoint_grouping}, namely using the Hough scores sampled at the keypoint positions as in \cite{papandreou2017towards} \vs the proposed Expected-OKS scoring of Eq.~\ref{eq:expected_oks}. We also compare the performance of the hard-NMS (using the OKS-based hard NMS scheme of \cite{papandreou2017towards} with threshold set to 0.5) \vs the proposed soft-NMS of Eq.~\ref{eq:instance_score_soft_nms}.

We show the results for the four alternative model configurations in Table~\ref{table:coco_keypoint_results_val_scoring_nms}. Both proposed components, Expected-OKS keypoint scoring and soft-NMS, bring significant improvements in AP over their alternatives from \cite{papandreou2017towards} and work well together.

\begin{table*}[h]
\centering
\caption{PersonLab performance on the COCO keypoints \textbf{val} split. Single-scale ResNet-101 model evaluation for different keypoint scoring and non-maximum suppression configurations. Largest image side is 1401 pixels and output stride is 8 pixels.}
\label{table:coco_keypoint_results_val_scoring_nms}
\scalebox{0.8}{
\begin{tabular}{cc|ccccc|ccccc}
Scoring & NMS &$AP$ & $AP^{.50}$ & $AP^{.75}$ & $AP^M$  & $AP^L$ & $AR$ & $AR^{.50}$ & $AR^{.75}$  & $AR^M$ & $AR^L$ \\
\hline
Hough \cite{papandreou2017towards} & hard & 0.632 & 0.838 & 0.693 & 0.593 & 0.698 & 0.682 & 0.862 & 0.733 & 0.635 & 0.751 \\
Expected-OKS                       & hard & 0.647 & 0.843 & 0.703 & 0.599 & 0.718 & 0.683 & 0.865 & 0.732 & 0.633 & 0.759 \\ \hline
Hough \cite{papandreou2017towards} & soft & 0.645 & 0.853 & 0.703 & 0.610 & 0.702 & 0.706 & 0.886 & 0.757 & 0.657 & 0.777 \\
\textbf{Expected-OKS}     & \textbf{soft} & 0.665 & 0.862 & 0.719 & 0.623 & 0.732 & 0.707 & 0.887 & 0.757 & 0.656 & 0.779 \\
\end{tabular}}
\end{table*}

\subsection{Ablation: Mid- and long-range offset refinement}

We examine the effect of mid- and long-range offset refinement on the quality of the keypoint and segmentation results.
For this purpose, we build a version of our model with offset refinement disabled during both training and evaluation.
Results on the COCO \textbf{val} split for the keypoints and segmentation tasks are shown in Tables~\ref{table:coco_keypoint_results_val_refinement} and~\ref{table:coco_segmentation_results_val_refinement}, respectively.
We see that offset refinement improves model keypoint AP by 3.3\% and segmentation AP by 2.2\%. In both cases, the largest improvement can be observed for large object instances, +5.4\% for keypoints and +9.1\% for segmentation, since large objects span a significant portion of the image for which accurate regression without refinement is challenging.

\begin{table*}[h]
\centering
\caption{PersonLab performance on the COCO keypoints \textbf{val} split. Single-scale ResNet-101 model evaluation without \vs with offset refinement. Largest image side is 1401 pixels and output stride is 8 pixels.}
\label{table:coco_keypoint_results_val_refinement}
\scalebox{0.8}{
\begin{tabular}{l|ccccc|ccccc}
        &$AP$ & $AP^{.50}$ & $AP^{.75}$ & $AP^M$  & $AP^L$ & $AR$ & $AR^{.50}$ & $AR^{.75}$  & $AR^M$ & $AR^L$ \\
\hline
Without offset refinement   & 0.632 & 0.856 & 0.689 & 0.603 & 0.678 & 0.679 & 0.883 & 0.736 & 0.639 & 0.735 \\
\textbf{With offset refinement} & 0.665 & 0.862 & 0.719 & 0.623 & 0.732 & 0.707 & 0.887 & 0.757 & 0.656 & 0.779 \\
\end{tabular}}
\end{table*}

\begin{table*}[h]
\centering
\caption{Performance on COCO Segmentation (Person category) \textbf{val} split. Single-scale ResNet-101 model evaluation without \vs with offset refinement. Inference with largest image side 1401 pixels, output stride 8 pixels, and 20 proposal budget.}
\label{table:coco_segmentation_results_val_refinement}
\scalebox{0.8}{
\begin{tabular}{l|cccccc|cccccc}
        & $AP$    & $AP^{50}$ & $AP^{75}$ & $AP^S$ & $AP^M$ & $AP^L$ & $AR^1$ & $AR^{10}$ & $AR^{20}$ & $AR^{S}$ & $AR^{M}$ & $AR^L$ \\ \hline
Without offset refinement   & 0.355 & 0.646 & 0.354 & 0.166 & 0.461 & 0.501 & 0.146 & 0.393 & 0.417 & 0.209 & 0.525 & 0.597 \\
\textbf{With offset refinement} & 0.382 & 0.661 & 0.397 & 0.164 & 0.476 & 0.592 & 0.162 & 0.416 & 0.439 & 0.204 & 0.532 & 0.681  \\
\end{tabular}}
\end{table*}

\subsection{Ablation: Small instance keypoint imputation in model training}

We examine the effect of imputing the keypoints of small COCO person instances and using them for model training.

When evaluating the model on the COCO keypoints task, keypoint imputation slightly decreases performance by 0.8\%, as seen in Table~\ref{table:coco_keypoint_results_val_imputation}. The reason is that the COCO keypoints evaluation protocol does not include the small person instances in the evaluation.

\begin{table*}[h]
\centering
\caption{PersonLab performance on the COCO keypoints \textbf{val} split. Single-scale ResNet-101 model evaluation when training without \vs with imputed small-instance keypoints. Inference with largest image side 1401 pixels and output stride 8 pixels.}
\label{table:coco_keypoint_results_val_imputation}
\scalebox{0.8}{
\begin{tabular}{l|ccccc|ccccc}
        &$AP$ & $AP^{.50}$ & $AP^{.75}$ & $AP^M$  & $AP^L$ & $AR$ & $AR^{.50}$ & $AR^{.75}$  & $AR^M$ & $AR^L$ \\
\hline
\textbf{Without imputation} & 0.665 & 0.862 & 0.719 & 0.623 & 0.732 & 0.707 & 0.887 & 0.757 & 0.656 & 0.779 \\
With imputation    & 0.657 & 0.864 & 0.718 & 0.617 & 0.723 & 0.705 & 0.891 & 0.760 & 0.655 & 0.776 \\
\end{tabular}}
\end{table*}

However, when evaluating the model on the COCO segmentation task, keypoint imputation significantly improves performance by 4.4\%, as shown in Table~\ref{table:coco_segmentation_results_val_imputation}. As expected, most of the performance improvement comes for small objects, whose AP more than doubles, increasing from 7.6\% to 16.4\%.

\begin{table*}[h]
\centering
\caption{Performance on COCO Segmentation (Person category) \textbf{val} split. Single-scale ResNet-101 model evaluation when training without \vs with imputed small-instance keypoints. Inference with largest image side 1401 pixels, output stride 8 pixels, and 20 proposal budget.}
\label{table:coco_segmentation_results_val_imputation}
\scalebox{0.8}{
\begin{tabular}{l|cccccc|cccccc}
        & $AP$    & $AP^{50}$ & $AP^{75}$ & $AP^S$ & $AP^M$ & $AP^L$ & $AR^1$ & $AR^{10}$ & $AR^{20}$ & $AR^{S}$ & $AR^{M}$ & $AR^L$ \\ \hline
Without imputation       & 0.338 & 0.560 & 0.368 & 0.076 & 0.459 & 0.591 & 0.156 & 0.370 & 0.383 & 0.080 & 0.514 & 0.680 \\
\textbf{With imputation} & 0.382 & 0.661 & 0.397 & 0.164 & 0.476 & 0.592 & 0.162 & 0.416 & 0.439 & 0.204 & 0.532 & 0.681 \\
\end{tabular}}
\end{table*}



\bibliographystyle{splncs}
\bibliography{biblio}
\end{document}